\documentclass[11pt, a4paper, nocopyright]{deepmind}

\usepackage[authoryear, sort&compress, round]{natbib}

\usepackage{placeins}
\usepackage[toc,page]{appendix}

\usepackage{nicefrac}       

\usepackage{boldline}       

\usepackage[font=small,labelfont=bf]{subcaption}
\usepackage[font=small]{caption}
\usepackage{verbatim}
\usepackage{xpatch}
\usepackage{setspace}
\usepackage{adjustbox}
\usepackage{amsmath, latexsym}
\usepackage{amsfonts}
\usepackage{amsthm}
\usepackage{xspace}
\usepackage{mathtools}

\usepackage{multirow}
\usepackage{arydshln}
\usepackage{rotating}
\usepackage{array}
\usepackage{mfirstuc}

\usepackage{amsmath, latexsym}
\usepackage{amsfonts}
\usepackage{mathtools}
\usepackage{amsthm}
\usepackage{dsfont}
\usepackage[dvipsnames]{xcolor}
\usepackage[colorinlistoftodos]{todonotes}
\usepackage{booktabs}
\usepackage{xfrac}
\usepackage{bbm}

\usepackage{xparse}
\usepackage{lipsum}
\usepackage{changepage}
\usepackage{enumitem}

\newcommand{\squishlist}{
   \begin{list}{$\bullet$}
    { \setlength{\itemsep}{0pt}      \setlength{\parsep}{3pt}
      \setlength{\topsep}{3pt}       \setlength{\partopsep}{0pt}
      \setlength{\leftmargin}{1.5em} \setlength{\labelwidth}{1em}
      \setlength{\labelsep}{0.5em} } }

\newcommand{\squishlisttwo}{
   \begin{list}{$\bullet$}
    { \setlength{\itemsep}{0pt}    \setlength{\parsep}{0pt}
      \setlength{\topsep}{0pt}     \setlength{\partopsep}{0pt}
      \setlength{\leftmargin}{2em} \setlength{\labelwidth}{1.5em}
      \setlength{\labelsep}{0.5em} } }

\newcommand{\squishend}{
    \end{list}  }

\newcolumntype{C}[1]{>{\centering\arraybackslash}m{#1}}
\newcolumntype{R}[1]{>{\raggedleft\arraybackslash}m{#1}}

\newcommand{\myvec}[1]{\boldsymbol{#1}}

\newcommand{\two}{(2)}

\newcommand{\va}{\myvec{a}}

\newcommand{\vr}{\myvec{r}}

\newcommand{\vu}{\myvec{u}}

\newcommand{\vx}{\myvec{x}}

\newcommand{\vz}{\myvec{z}}

\newcommand{\real}{\mathbb{R}}

\newcommand{\union}{\cup}

\newcommand{\calA}{{\cal A}}

\newcommand{\calR}{{\cal R}}

\DeclareMathAlphabet{\mathpzc}{OT1}{pzc}{m}{n}

\usepackage{makecell}
\usepackage{fancyvrb}

\RequirePackage{algorithm}
\RequirePackage{algorithmic}
\renewcommand{\algorithmiccomment}[1]{\hfill$\blacktriangleright$ #1}
\renewcommand{\algorithmicrequire}{\textbf{Input:}}
\renewcommand{\algorithmicensure}{\textbf{Output:}}

\correspondingauthor{$\lbrace$lmatthey, lerchner$\rbrace$@google.com}


\reportnumber{} 

\title{COBRA: Data-Efficient Model-Based RL through Unsupervised Object Discovery and Curiosity-Driven Exploration}

\author[*,1]{Nicholas Watters}
\author[*,1]{Loic Matthey}
\author[1]{Matko Bo\v snjak}
\author[1]{Christopher P. Burgess}
\author[1]{Alexander Lerchner}

\affil[*]{Equal contribution}
\affil[1]{DeepMind}

\renewcommand{\today}{2019-05-22}

\begin{abstract}
Data efficiency and robustness to task-irrelevant perturbations are long-standing challenges for deep reinforcement learning algorithms.
Here we introduce a modular approach to addressing these challenges in a continuous control environment, without using hand-crafted or supervised information.
Our Curious Object-Based seaRch Agent (COBRA) uses task-free intrinsically motivated exploration and unsupervised learning to build object-based models of its environment and action space.
Subsequently, it can learn a variety of tasks through model-based search in very few steps and excel on structured hold-out tests of policy robustness.
\end{abstract}

\begin{document}
\maketitle
\balance

\vskip 0.3in

\vspace*{-10pt}
\section{Introduction}\label{S:intro}

Recent advances in deep reinforcement learning (RL) have shown remarkable success on challenging tasks \citep{dqn2015, alphago, openaimanipulation}.
However, data efficiency and robustness to new contexts remain persistent challenges for deep RL algorithms, especially when the goal is for agents to learn practical tasks with limited supervision.
Drawing inspiration from self-supervised ``play'' in human development \citep{gopnik1999, settles2011}, we introduce an agent that learns object-centric representations of its environment without supervision and subsequently harnesses these to learn policies efficiency and robustly.

Our agent, which we call \textbf{C}urious \textbf{O}bject-\textbf{B}ased sea\textbf{R}ch \textbf{A}gent (\textbf{COBRA}), brings together three key ingredients: (i) learning representations of the world in terms of objects, (ii) curiosity-driven exploration, and (iii) model based RL.
The benefits of this synthesis are data efficiency and policy robustness.
To put this into practice, we introduce the following technical contributions:
\begin{itemize}
    \item A method for learning action-conditioned dynamics over slot-structured object-centric representations that requires no supervision and is trained from raw pixels.
    \item A method for learning a distribution over a multi-dimensional continuous action space. This learned distribution can be sampled efficiently.
    \item An integrated continuous control agent architecture that combines unsupervised learning, adversarial learning through exploration, and model-based RL.
\end{itemize}

COBRA is trained in two phases.
During the initial \textbf{exploration phase} it explores its environment, in which it can move objects freely with a continuous action space but is not rewarded for its actions.
In this phase, it learns how to see, predict, and act in a task-free setting.
It uses these capacities during a subsequent \textbf{task phase}, in which it is trained through model-based RL and quickly learns to solve tasks.

\section{Related Work}\label{S:related}

Our work builds upon three areas of prior research: Curiosity-driven exploration, object-oriented RL and model-based RL.

\paragraph{Curiosity-driven exploration} was first proposed in \citep{schmidhuber1990making, schmidhuber1991possibility} and expanded upon in \citep{schmidhuber2010formal}.
These works introduce the notion of learning a ``world model'', which can be used to plan or explore given a ``curiosity'' measure.
More recently, some models \citep{haber2018learning, laversanne2018curiosity} have shown interesting exploration behaviors emerge through task-free curiosity-driven exploration in agents.
Others have shown benefits of intrinsic motivation for learning RL tasks \citep{kulkarni2016, eysenbach2018diversity, pathak2017curiosity, song2019}, though the connections to policy robustness and object-based representations are not explored in these works.

\paragraph{Object-Oriented RL}
Proposed as an alternative to classical MDPs, object-oriented RL~\citep{diuk2008object} promises data efficiency and generalization by leveraging representations of objects and their relations \citep{keramati2018}.
Intuitively, representations factored in terms of objects support structured reasoning and facilitate solving many tasks.
Extensions of these approaches can be compatible with hierarchical RL \citep{roderick2017, vezhnevets2017feudal}.
However, most existing approaches require hand-crafting object representations or their relations \citep{diuk2008object, cobo2013object, garnelo2016towards, vicarious2019}.
In contrast, our approach addresses the core problem of automatically discovering these from data without hand-crafting or supervision.

\paragraph{Model-based RL}
The promises of learning and leveraging a model of the environment have been discussed before \citep{sutton2018reinforcement}, yet doing so in practice for complex environments remains a challenge \citep{deisenroth2011pilco}.
Recent works have made progress in this regard \citep{finn2016deep, ha2018world, zhang2018solar, kurutach2018learning, kaiser2019model, sodhani2019}, but model-free alternatives are still hard to beat in terms of asymptotic performance.
Concurrent work by \citet{hafner2018learning} combines intrinsic motivation with model-based RL on continuous control tasks.
Our model differs in that it learns to infer a more structured latent representation by decomposing scenes into their constituent objects, which we hypothesized will help us scale up to more complex situations.
However, our current policy learning approach is more limited, and hence a careful comparison with the \citet{hafner2018learning} model would be interesting for future work.

\section{The COBRA model}\label{S:model}

\begin{figure*}[t]
  \centering
  \includegraphics[width=1\linewidth]{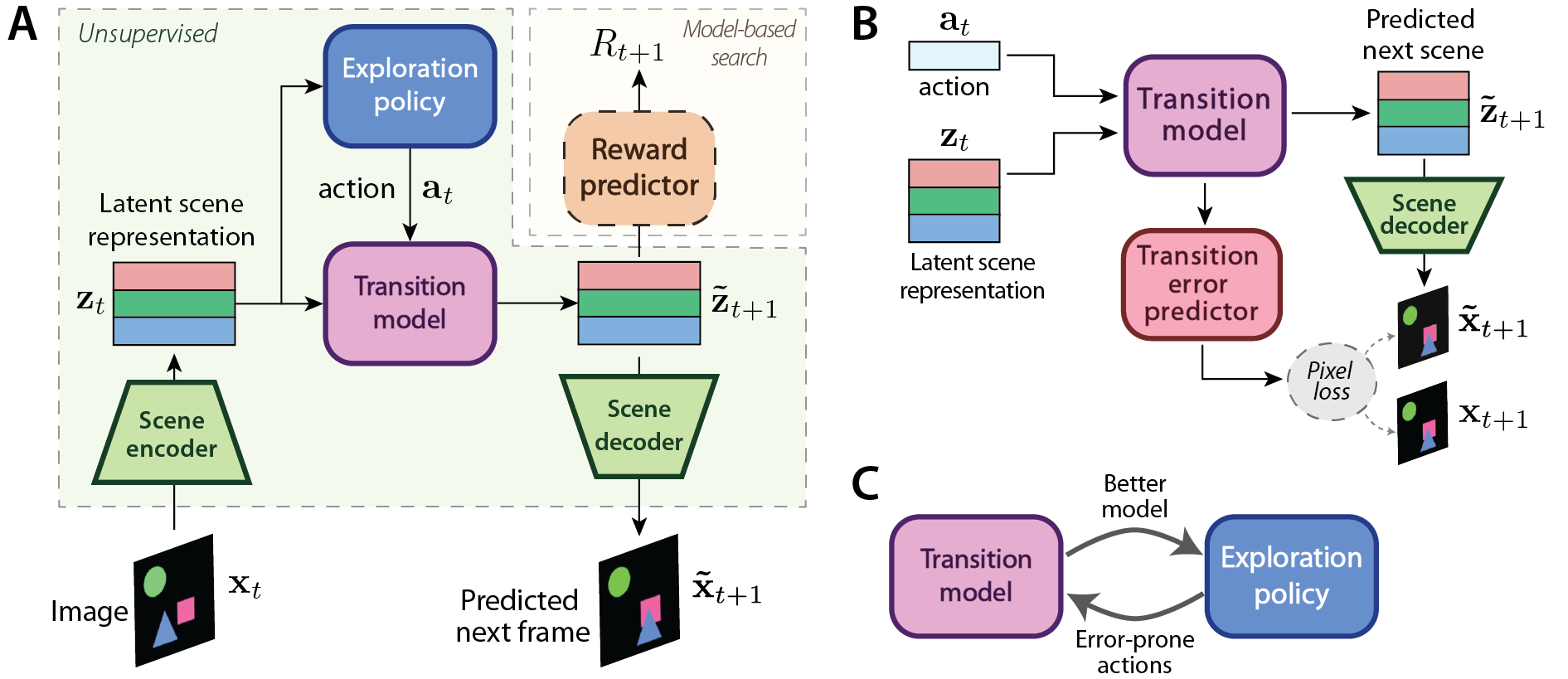}
  \caption{\textbf{COBRA model schematic}\\
  \textbf{A.} Entire model. The vision module (scene encoder and decoder), transition model, and exploration policy are all trained in a pure exploration phase with no reward. \textbf{B.} Transition model architecture. An action-conditioned slot-wise MLP learns one-step future-prediction. This is trained by applying the scene decoder to $\tilde \vz_{t+1}$, through which gradients from a pixel loss are passed. An auxiliary transition error prediction provides a more direct path to the pixel loss and makes adversarial training with the exploration policy more efficient. \textbf{C.} Adversarial training of transition model and exploration policy through which the behavior of moving objects emerges.
  }
  \label{fig:model_schematic}
  \vspace*{-1em}
\end{figure*}

Our model (Figure~\ref{fig:model_schematic}) consists of four components: The \textbf{vision module}, \textbf{transition model}, and \textbf{exploration policy} are trained during an unsupervised \textit{exploration phase} without rewards, while the \textbf{reward predictor} is trained during a subsequent \textit{task phase} (with reward).

To present the architectures of these components, we must first briefly discuss the environment:
We are interested in environments containing objects that can be manipulated, in which agents can learn purposeful exploration without reward and a diverse distribution of tasks can be defined.
In all of our experiments we use a 2-dimensional virtual ``touch-screen'' environment that contains objects with configurable shape, position, and color.
The agent can move any visible object by clicking on the object and clicking in a direction for the object to move.
Hence the action space is continuous and 4-dimensional, namely a pair of clicks.
If the first click does not land in an object then nothing will move, regardless of the second click.
Despite its apparent simplicity, this environment supports highly diverse tasks that can challenge existing SOTA agents.
See Section \ref{S:env_tasks} for details about our environment and tasks.
We have open-sourced this environment \footnote{Available here: \url{https://github.com/deepmind/spriteworld}}, calling it ``Spriteworld'' \citep{spriteworld}.

\subsection{Unsupervised exploration phase}\label{SS:unsupervised_exploration_phase}

We want our agent to learn how its environment ``works'' in a task-free \textit{exploration phase}, where no extrinsic reward signals are available.
If this is done well, then the knowledge thus gained should be useful and generalizable for subsequent tasks.
Specifically, in the exploration phase COBRA learns representations of objects, dynamics, and its own action space, which it can subsequently use for any task.
For the exploration phase we let COBRA explore our Spriteworld environment in short episodes.
Each episode is initialized with 1-7 objects with randomly sampled shape, position, orientation and color.

\paragraph{Vision Module} We use MONet \citep{burgess2019} to learn a static scene representation.
MONet is an auto-regressive VAE that learns to decompose a scene into entities, such as objects, without supervision.
We can view it as having an encoder $V_{enc}$ and a decoder $V_{dec}$.
The encoder maps an image $\vx$ to a tensor $V_{enc}(\vx) = \vz \in \real^{K \times M}$ representing the full scene.
This representation consists of $K$ entities, each encoded by the mean of MONet's latent distribution of dimension $M$.
We hereafter call each row $k$ of the scene representation $\vz$ a ``slot.''
The decoder maps this tensor $\vz$ to a reconstructed image $V_{dec}(\vz) = \tilde{\vx}$.

\begin{figure}[t!]
  \centering
  \includegraphics[width=0.99\linewidth]{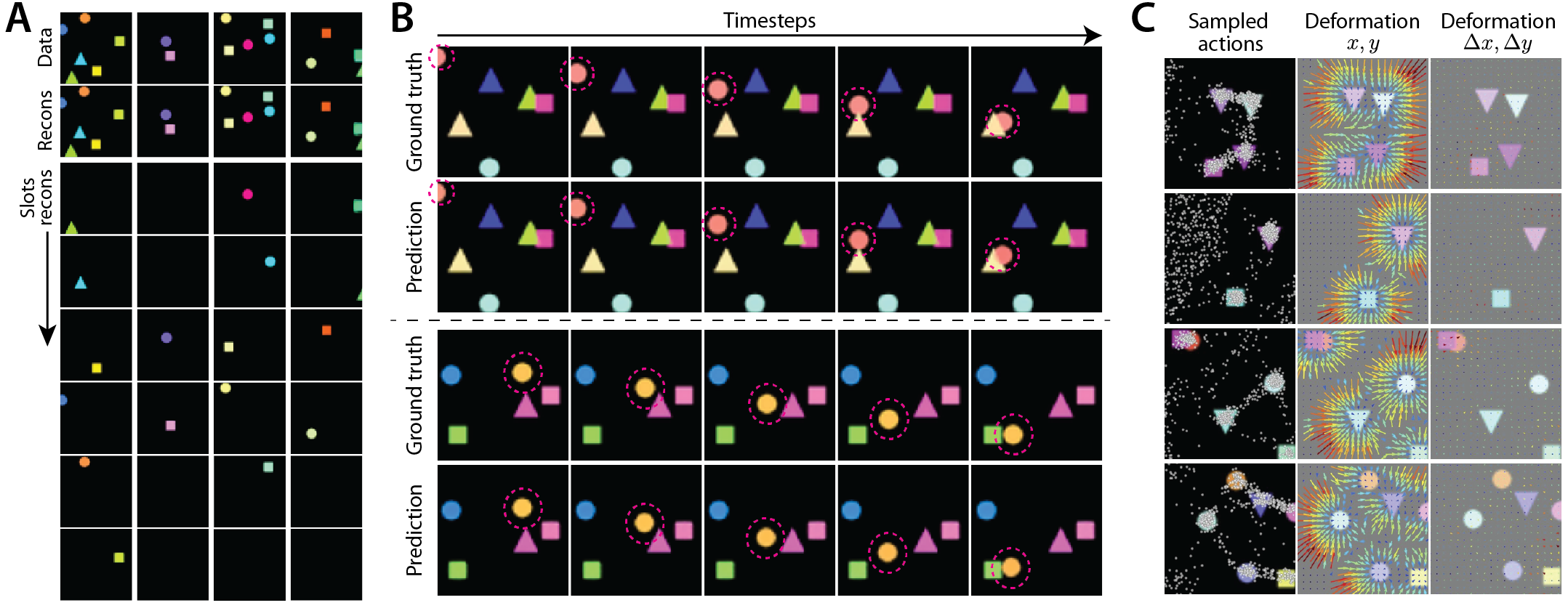}
  \caption{\textbf{Scene decomposition, transition model, and exploration policy results}\\
  \textbf{A.} Vision module (MONet) decomposing Spriteworld scenes into objects (one column per sample scene).
  \textit{(First row)} Data samples.
  \textit{(Second row)} Reconstruction of the full scene.
  \textit{(Other rows)} Individual reconstructions of each slot in the learned scene representation.
  Some slots are decoded as blank images by the decoder.
  \textbf{B.} Rollouts of transition model, treated as an RNN, compared to ground truth on two scenes.
  In each scene, one single item (indicated by dotted circle) is being moved along a line in multiple steps.
  \textbf{C.} Exploration policy.
  \textit{(Left)} Position click component of random samples from the trained exploration policy, which learns to click on (and hence move) objects.
  \textit{(Middle)} Slice through the first two dimensions (position click) of the action sampler's quantile function, showing deformations applied on a grid of first clicks $\in [0, 1]^2$ with randomized second click.
  \textit{(Right)} Slice through second two dimensions (motion click). There is virtually no deformation, indicating the exploration policy learns to sample motions randomly.
  }\label{fig:scene_action_transition}
\end{figure}

Critically, MONet learns to put each object from a scene into a different slot.
Moreover, the MONet architecture ensures that each slot has the same representational semantics, so there is a common learned latent space to which each object is mapped.
Each slot represents meaningful properties of the objects, such as position, angle, color and shape.
See Figure~\ref{fig:supp_disentangling} in Appendix~\ref{S:supp_disentangling} for an example.
MONet can handle scene datasets with a variable number of objects by letting some slots' codes lie in a region of latent space that represents no object.
See Figure \ref{fig:scene_action_transition}-A for scene decomposition results, indicating that MONet successfully learns to accurately represent Spriteworld scenes in terms of objects.

While our Spriteworld environment is visually quite simple, in more complex environments MONet's slots learn to represent other elements of a scene (e.g. backgrounds or walls in 3D scenes) and its latent space can capture more complex visual properties (e.g. surface glossiness).
See \citep{burgess2019} for examples.

\paragraph{Transition model} We introduce a method that maps an action $\va_t$ and a scene representation $\vz_t$ at time step $t$ to a predicted next-step scene representation $\tilde \vz_{t+1}$.
This model applies a shared MLP $T_{net}$ slot-wise to the concatenations $[\vz_t^k; \va_t]$ of each slot $\vz_t^k$ with the action $\va_t$.
This is sufficient to predict $\vz_{t+1}$, because in Spriteworld there are no physical interactions between objects and any action will affect at most one object.
To train this transition model, we cannot easily use a loss in representation space, because the encoding $\vz_{t+1} = V_{enc}(\vx_{t + 1})$ of image $\vx_{t + 1}$ may have a different ordering of its slots than $\vz_t = V_{enc}(\vx_t)$ and solving the resulting matching problem in a general, robust way is non-trivial.
Instead, we circumvent this problem by applying the visual decoder (through which gradients can be passed) to $\tilde \vz_{t+1}$ and using a pixel loss in image space:
\begin{align}
    \tilde \vz_{t+1} = T(\vz_t, \va_t) &= [T_{net}(\vz_t^0, \va_t), ..., T_{net}(\vz_t^K, \va_t)]\nonumber\\
    Loss_{T}(\vz_t, \va_t) &= ||V_{dec}( T(\vz_{t}, \va_t)) - \vx_{t + 1}||^2\nonumber
\end{align}
Additionally, we train an extra network $L_T(\vz, \va)$ to predict the output of the pixel loss, which we use as a measure of curiosity when training the exploration policy, as we found this to work better and be more stable than using the pixel loss directly.
See Appendix~\ref{S:supp_things_tried} for alternative transition models considered, including those that do not use pixel loss.

See Figure \ref{fig:scene_action_transition}-B for examples of transition model rollouts after training in tandem with the exploration policy (covered below).

\begin{figure}[t!]
  \centering
  \includegraphics[width=0.55\linewidth]{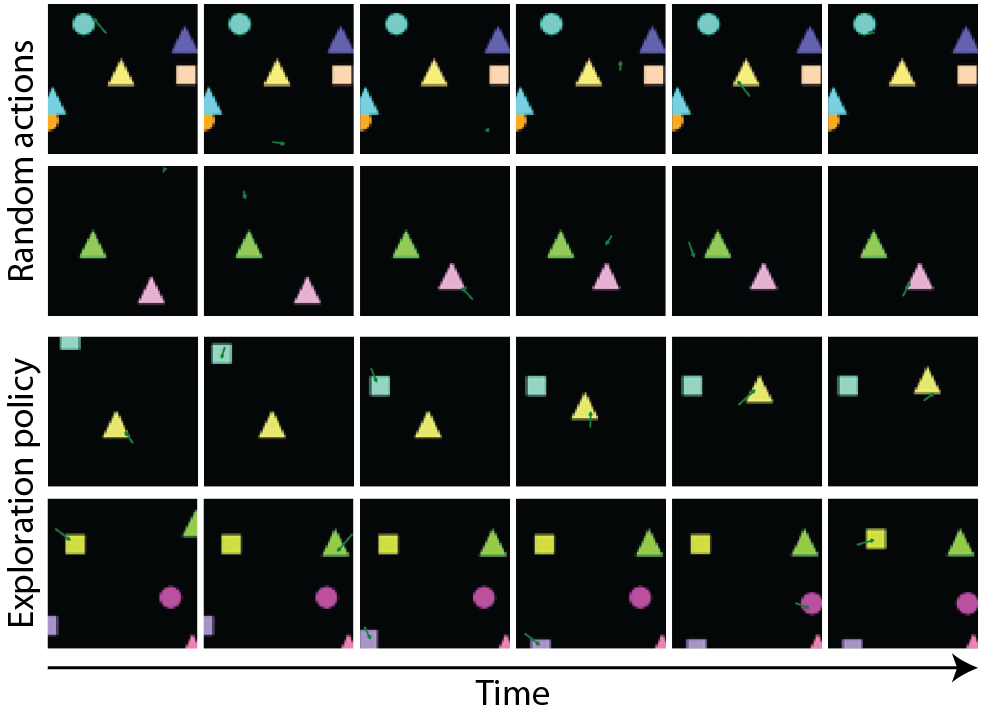}
  \caption{
  \textbf{Random policy and exploration policy}\\
  Observations and actions taken by an agent during the unsupervised exploratory phase. 
  Actions are shown with small green arrows.
  \textit{(Top)} Random agent, which rarely moves any object, provides a bad source of data for the transition model.
  \textit{(Bottom)} trained exploration policy, which frequently moves objects, provides a good source of data for the transition model.
  }\label{fig:supp_exploration_behaviour}
\end{figure}

\paragraph{Exploration Policy} In many environments a uniformly random policy is insufficient to produce action and observation sequences representative of those useful for tasks.
Consequently, in order for a agent to learn through pure exploration a transition model that transfers accurately to a task-driven setting, it should not take uniformly random actions.
In Spriteworld, this manifests itself in the sparseness of the action space.
If an agent were to take uniformly random actions, it would rarely move any object because the agent must click within an object to move it and the objects only take up a small portion of the image ($\approx 1.7 \%$ per object).
This is shown in Figure~\ref{fig:supp_exploration_behaviour} (top) and Appendix~\ref{S:supp_agent_videos}, and our transition model does not get enough examples of objects moving to be trained well in this condition.
Hence we need our agent to learn in the exploration phase a policy that clicks on and moves objects more frequently.

Our approach is to train the transition model adversarially with an exploration policy that learns to take actions on which the transition model has a high error.
Such difficult-to-predict actions should be those that move objects (given that others would leave the scene unchanged).
In this way the exploration policy and the transition model learn together in a virtuous cycle.
This approach is a form of curiosity-driven exploration, as previously described in both the psychology \citep{gopnik1999} and the reinforcement learning literature \citep{schmidhuber1990line, schmidhuber1990making, pathak2017curiosity}.
We apply this adversarial training of the transition model and exploration policy in an exploration environment with randomly generated objects (see Appendix~\ref{S:supp_training}).

To put this idea into practice, we must learn a distribution over the 4-dimensional action space from which we can sample.
To do this, we take inspiration from distributional RL \citep{distributional_rl, dabney2017quantile, dabney2018quantile} and learn an approximate quantile function that maps uniform samples $\vu 
\in [0, 1]^4$ to non-uniform samples in our action space $\va \in [0, 1]^4$.
This can be thought of as a perturbation $D$ of each point in $[0, 1]^4$, parameterized by an MLP $D_{net}$.
We train $D_{net}$ to maximize the predicted error $L_T$ of the transition model's prediction of action $\va$, subject to an $L_1$ regularization on the perturbation magnitude. Specifically, given $\vu\in [0, 1]^4$, the exploration policy is trained as:
\begin{align}
    D(\vu, \vz) &= \vu + D_{net}(\vu, \vz) = \va \nonumber\\
    Loss(\vu, \vz) &= |D_{net}(\vu, \vz)| - L_T(\vz, \va)\nonumber
\end{align}
Note that this method does not pressure $D$ to exactly be a quantile function.
Instead it is incentivized to learn a discontinuous approximation of the quantile function.
We tried various approaches to parameterizing $D$ (see Appendix~\ref{S:supp_things_tried}) and found this one to work best.

While the vision module, transition model, and exploration policy can in principle all be trained simultaneously, in practice for the sake of simplicity in this current work we pre-trained the vision module on frames generated by taking random actions in the exploration environment.
We then reloaded this vision module with frozen variables while training the transition model and exploration policy in tandem in the active adversarial exploration phase just described.
See Appendices~\ref{S:supp_model}-\ref{S:supp_training} for architecture and training details.

See Figure \ref{fig:scene_action_transition}-C for examples of samples from the trained exploration policy and a visualization of its deformation.
Images of the exploration policy in action can be seen in Figure~\ref{fig:supp_exploration_behaviour} (bottom), and see Appendix~\ref{S:supp_agent_videos} for videos.

\subsection{Task phase}\label{SS:task_phase}

Once the exploration phase is complete, COBRA enters the \textit{task phase}, in which it is trained to solve several tasks independently.
We freeze the trained vision module, transition model, and exploration policy, training only a separate reward predictor independently for each task.
See Section~\ref{S:env_tasks} for task details.

\paragraph{Reward Predictor}

For each task we train a model-based agent that uses the components learned in the exploration phase and only learns a reward predictor.
The reward predictor is trained from a replay buffer of (\texttt{observation}, \texttt{reward}) pairs, see Algorithm~\ref{alg:agent}.

Our agent acts with a simple 1-step search, sampling a batch of actions from the exploration policy, rolling them out for one step with the transition model, evaluating the predicted states with the reward predictor and selecting the best action (with epsilon-greedy).
This is effectively a myopic 1-step Model Predictive Control (MPC) policy.
It is sufficient to act optimally given the dense rewards in our tasks, but could readily be extended to multi-step MPC, other existing planning algorithms as in \cite{hafner2018learning} or Monte-Carlo Tree Search \citep{alphago}.
Also, see Appendix \ref{S:supp_sparse_rewards} for an extension that works with sparse rewards.

\begin{figure}[t!]
  \centering
  \begin{minipage}{.7\linewidth}
    \begin{algorithm}[H]
    \caption{Task phase agent training}
    \label{alg:agent}
    \begin{algorithmic}
        \REQUIRE Branching factor $B$, Training factor $N$
        \ENSURE Trained reward predictor $R_\theta$
        \STATE $\calR = \varnothing$
        \COMMENT{initialize replay buffer}
        \FOR{$t=1$ {\bfseries to} \texttt{num\_steps}}
        \STATE $\vx_t, r_t = \texttt{environment.obs\_reward()}$
        \STATE $\vz_t = V_{enc}(\vx_t)$
        \COMMENT{scene representation}
        \STATE $\calR \leftarrow \calR \union (\vz_t, r_t)$
        \STATE $\calA = \varnothing$
        \COMMENT{actions and predicted rewards}
        \FOR{$b=1$ {\bfseries to} $B$} 
            \STATE $\vu_b \sim \text{Uniform}([0, 1]^4)$
            \STATE $\va_b = D(\vz_t, \vu_b)$
            \COMMENT{exploration policy action}
            \STATE $\tilde \vz_{t+1} = T(\vz_t, \va_b)$
            \COMMENT{transition model}
            \STATE $\tilde r_b = R_\theta(\tilde \vz_{t+1})$
            \COMMENT{reward predictions}
            \STATE $\calA \leftarrow \calA \union (\va_b, r_b)$
        \ENDFOR
        \STATE $\va = \text{epsilon-greedy}(\max_{\tilde r}(\calA))$
        \COMMENT{best action}
        \STATE $\texttt{environment.step}(\va)$
        \FOR{$i=1$ {\bfseries to} $N$} 
        \STATE $\vz_i, \vr_i \sim \calR$
        \COMMENT{sample minibatch}
        \STATE $\theta \leftarrow \theta - \partial_{\theta} ||R_\theta(\vz_i) - \vr_i||^2$
        \COMMENT{train reward predictor}
        \ENDFOR
        \ENDFOR
    \end{algorithmic}
    \end{algorithm}
  \end{minipage}
\end{figure}

\section{Environment and Tasks}\label{S:env_tasks}

Our Spriteworld environment (\citet{spriteworld}, open-sourced at \url{https://github.com/deepmind/spriteworld}) is a 2-dimensional square arena with a variable number of colored sprites, freely placed and rendered with occlusion but no collisions (see Figures~\ref{fig:model_schematic}-\ref{fig:scene_action_transition}).
Agents use a continuous 4-dimensional click-and-push action space, where an action is a point in the hypercube $[0, 1]^4$.
The first two coordinates of an action are a "position click" $(x, y)$ and the second two are a "motion click" $(\Delta x, \Delta y)$.
If the position click, treated as a point within the environment, falls within an object's boundaries, then that object will take a small step in the direction specified by the motion click.
If the position click does not fall within an object then no object moves, regardless of the motion click.

While this environment looks visually simple, it has some important features:
\begin{itemize}
    \item The multi-object arena reflects the compositionality of the real world, with cluttered scenes of objects that can share features yet move independently. This also provides ways to test robustness to task-irrelevant features/objects and combinatorial generalization.
    \item The structure of the continuous click-and-push action space reflects the structure of space and motion in the world. It also allows the agent to move any visible object in any direction.
    \item The notion of an object is not provided in any privileged way (e.g. no object-specific components of the action space) and can be fully discovered by agents.
\end{itemize}

Furthermore, difficult tasks can be designed in this environment that challenge state-of-the-art continuous control agents.
We consider a suite of six tasks, grouped into three categories.
For each task we have a held-out extrapolation set to test the agent's policy robustness to task-irrelevant properties:
\begin{itemize}
    \item \textbf{Goal-Finding.}
    The agent must bring a set of target objects (identifiable by some feature, e.g. "green") to a hidden location on the screen, ignoring distractor objects (e.g. those that are not green)
    This goal location is fixed across episodes.
    To examine robustness, we test extrapolation over the number of targets, number of distractors, and task-irrelevant target features.
    \item \textbf{Sorting.}
    The agent must bring each object to a goal location based on the object's color.
    To examine robustness, we test extrapolation to an object combination that was not seen during training time (e.g. trained on \{blue, red\} pairs and \{red, green\} pairs, tested on \{blue, green\} object pairs).
    This tests whether the agent can learn to factorize and compose independent goals.
    \item \textbf{Clustering.}
    The agent must arrange the objects in clusters according to their color. We test robustness to colors different from those used for training.
\end{itemize}
See Figure~\ref{fig:agent_behaviour} for a visualization of our agent solving them (note that the targets are only shown for visualization purposes and are not provided to the agent).
Some of our robustness tests are rather mild (e.g. robustness to different shapes when only color matters), while others are more demanding (e.g. increasing the number of target objects, or clustering novel color combinations).
However, we believe all are intuitively reasonable behaviors to desire.
See Appendix~\ref{S:supp_environment} for more details about Spriteworld, tasks, and reward functions.

\section{Results}\label{S:results}

We now show results on solving the tasks described above.
As explained before, we are interested both in data efficiency (in terms of rewarded environment steps), as well as the robustness of the policies to task-irrelevant perturbations.

As our Spriteworld environment uses sparse continuous actions, finding appropriate baselines is challenging.
We compare our agent to two baselines:
\begin{itemize}
    \item[i)] \emph{MPO raw}:
    Maximum a Posteriori Policy Optimization (MPO)~\citep{abdolmaleki2018maximum}, a state-of-the-art model-free continuous control algorithm known for its data-efficiency.
    \item[ii)] \emph{MPO handheld}:
    MPO endowed with our agent's vision module and exploration policy.
    This agent applies the vision model's encoder to its image observations received from the environment and applies the exploration policy's deformation to its actions before passing them to the environment.
\end{itemize}

\begin{figure}[t!]
  \centering
  \includegraphics[width=1.\linewidth]{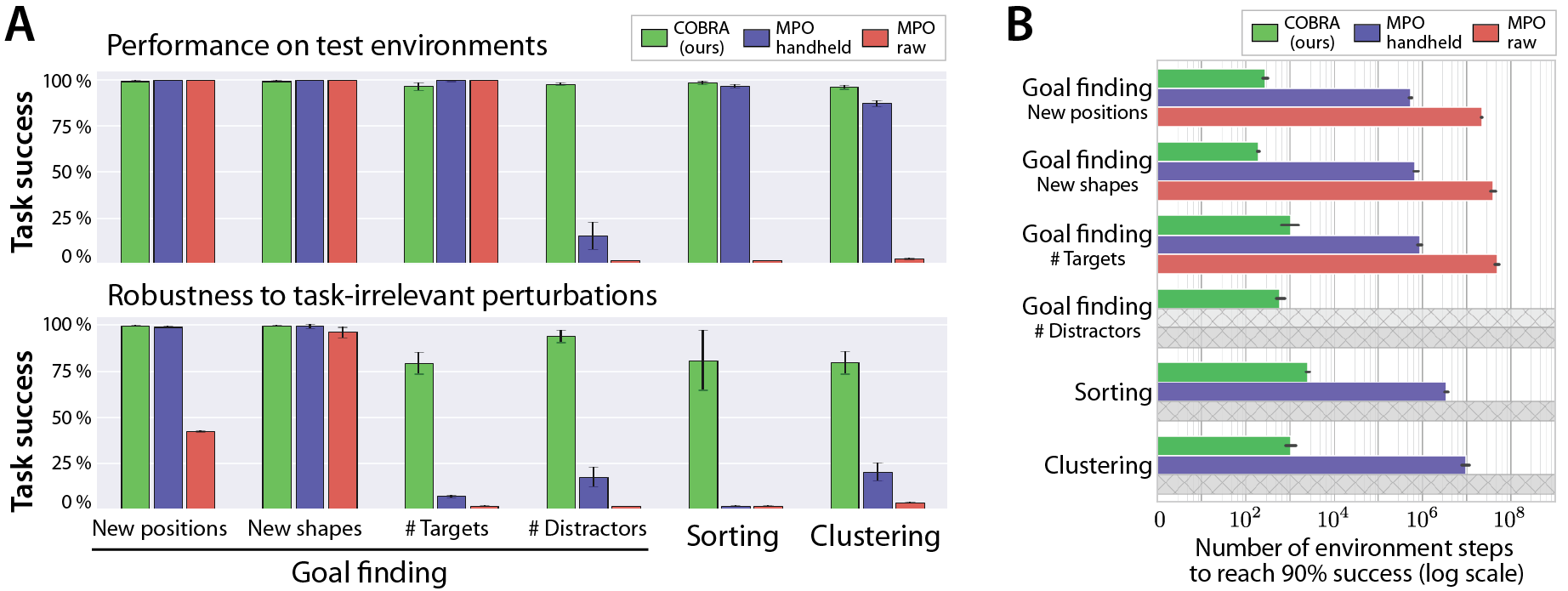}
  \caption{\textbf{Performance, Robustness, and Data Efficiency}\\
  \textbf{A.} Performance and robustness of agents after training until convergence.
  Top row shows test-time performance of agents on random environments sampled from the training distribution (higher is better).
  Bottom row shows robustness tests to out-of-distribution task-irrelevant environment perturbations (see main text for details).
  \textbf{B.} Data efficiency (lower is better). 
  Computed as smallest number of on-task environment steps needed to reach and sustain $90\%$ average performance over 30 consecutive episodes.
  The corresponding number of episodes varies depending on task and agent performance, but for COBRA ranges from $\sim 15$ (Goal finding new shapes) to $\sim 600$ (sorting).
  Gray bars indicate no agent reached $90\%$ performance.
  }\label{fig:agent_results}
\end{figure}

Because MPO is model-free, the \emph{MPO raw} agent must be entirely re-trained for all tasks. This makes comparing data efficiency to our model difficult because our model's unsupervised exploration phase only needs to be trained once and can then be used for any task.
The data-efficiency comparison with \emph{MPO handheld} is fairer because \emph{MPO handheld} re-uses the same amount of pre-training as our agent.
We posit that our model's extreme on-task data efficiency (see log-scale in Figure \ref{fig:agent_results}-B) justifies our training paradigm, and note that even our agent's exploration phase is more data efficient than \emph{MPO raw} on any of the tasks.
COBRA learns to solve most of our tasks using only 100-1000 rewarded environment steps, compared to $10^6 - 10^8$ for all MPO baselines.
We can expect these gains to especially add up when the number of tasks to be achieved in this environment increases, as we effectively amortize the cost of our pretraining across all future tasks.

\begin{figure}[t!]
  \centering
  \includegraphics[width=0.62\textwidth]{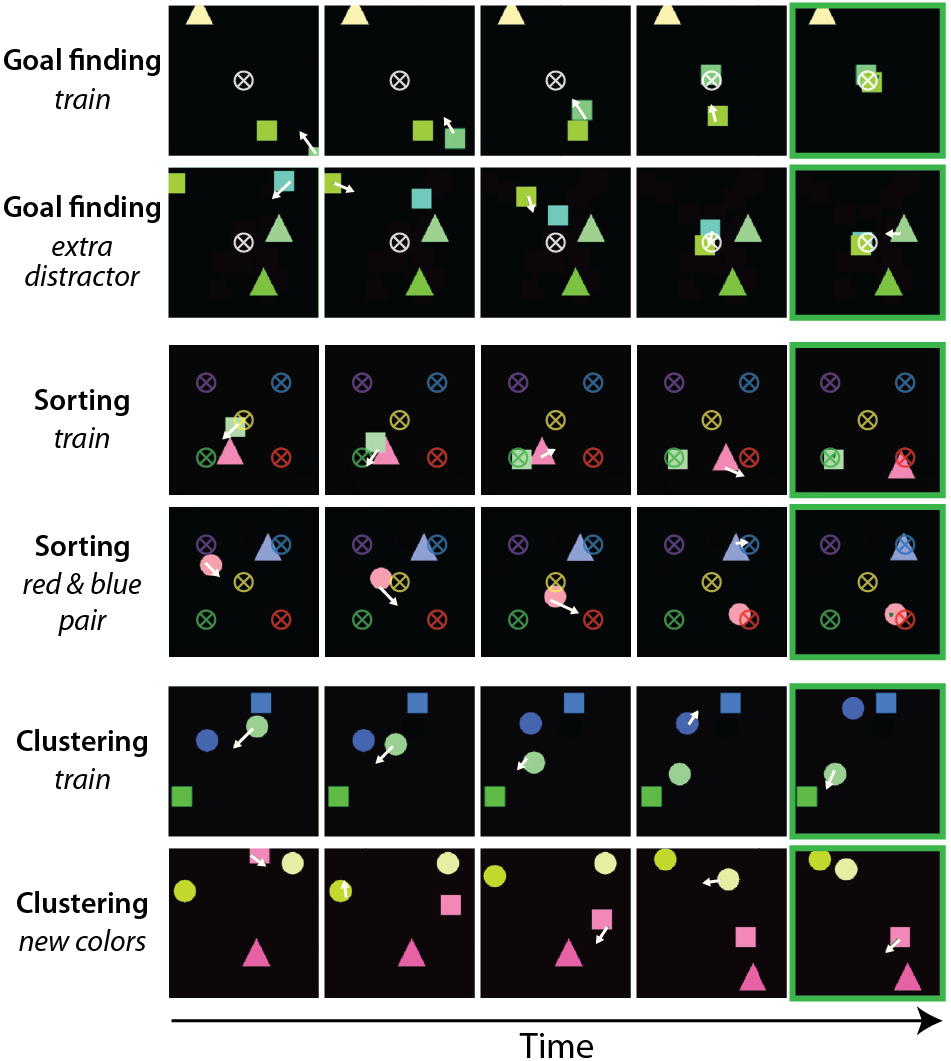}
  \caption{\textbf{COBRA solving our tasks}\\
  Demonstration of a trained COBRA agent solving different tasks and its behaviour on robustness tests.
  Agent actions are shown with white arrows, and target goals are shown with crossed circles.
  These targets are only shown for visualization purposes and are not provided to the agent.
  See Appendix~\ref{S:supp_agent_videos} for links to videos. Only five steps are displayed for each episode.
  \textbf{(Top)} Solving a ``Goal finding'' task.
  Having been trained only with a single distractor, COBRA is robust to the addition of a second distractor at test time.
  \textbf{(Middle)} Solving a ``Sorting by color'' task.
  COBRA has been trained to bring objects to different goals depending on their colours, seeing all pairs of colors except (blue, red).
  It is robust to testing on this held out combination of objects and successfully brings them to their targets.
  \textbf{(Bottom)} Solving a ``Cluster by color'' task.
  Having been trained only on clustering green/blue objects, COBRA successfully extrapolates its reward predictor, and hence its policy, to clustering red/yellow objects.
  }\label{fig:agent_behaviour}
\end{figure}

Our model performs on the robustness tests (Figure \ref{fig:agent_results}-A), far exceeding the baselines on many of them.
We presume this arises because the task-specific portion of our model is minimal, reducing over-fitting.
Specifically, our model learns only a reward function, not a policy network.
In general, we conjecture that learning goals and using a model to plan a policy results in more robustness than can be achieved with model-free RL.
Example behaviors of our agent solving tasks are shown in Figure~\ref{fig:agent_behaviour} (see also Appendix~\ref{S:supp_agent_videos} for links to videos).

\section{Discussion}\label{S:discussion}

We have introduced COBRA, which to our knowledge is the first agent to combine unsupervised learning of object-centric representations, curiosity-driven exploration, and model-based RL all together into one architecture. We demonstrated how this approach can be used to both achieve very high data-efficiency when learning tasks and yield policies that are robust to task-irrelevant perturbations.

We introduced the Spriteworld environment, which we used for all of our experiments. Spriteworld has a continuous action space where the agent can move any object, yet does not hand-craft the notion of object in any way.
We considered instead using a discretized version of the touch-screen action space consisting of a fine mesh in $[0, 1]^4$.
While this would have allowed us to compare to numerous model-based discrete-action-space baselines, we found in early experiments that it was difficult to get any agent to train with such a large discrete action space.
Furthermore, we believe that the continuous version provides a more realistic geometric relationship between action space and environment.
We also considered using the action space of \citet{haber2018learning}, which has a distinct component controlling each object in the scene (according to an arbitrary ordering).
While that would have made exploration much easier, we believe it builds in the notion of an object in a way that circumvents the important problem of object discovery and is unlikely to scale to large numbers of objects.

Our current version of the transition model does not take into account interactions between objects, though we think this limitation can be readily overcome by incorporating a GraphNet \citep{graph_nets}.
Furthermore, it might seem that training the transition model using a pixel loss could give rise to problems for small objects.
However, as we do not update the variables of MONet based on the transition model's pixel loss, our transition model does not suffer from this issue.
One limitation of our transition model is its lack of memory.
While this is fine in a fully observable environment like Spriteworld, in more complex environments we would like a transition model that remembers out-of-view objects.
Exploring this would be an interesting direction of future work.

Our exploration policy is a form of distributional RL,  \citep{distributional_rl, dabney2018quantile, dabney2017quantile}, though we found that the Huber quantile regression loss used in existing distributional RL work scaled poorly with dimension and was unable to learn the sparse multimodal 4-dimensional distributions needed by COBRA. This motivated our alternative deformation-based approximate quantile loss, which we found to train quite robustly.
Note, however, that our method is not pressured to learn a smooth quantile function, and is better thought of as learning basins of attraction around high-value (i.e. high-curiosity) actions.
While we found this to scale well with dimensionality, it has no pressure to force the basins of attraction to all have the same volume, hence in Spriteworld it prefers moving objects surrounded by empty space to moving objects next to a boundary or other object.
We did not find this to impact agent performance, though might be a concern in more cluttered environments.

Our use of transition model error as a metric of curiosity is quite simple. While it works on Spriteworld, it would not be suitable for environments with unpredictable dynamics.
The Intrinsic Curiosity Module of \citet{pathak2017curiosity} is complementary to our model and would be a natural way to address this limitation.
The approach in \citep{laversanne2018curiosity}, which samples goals according to predicted policy improvement, is also appealing, though parameterizing the goal space is challenging in environments with a variable number of objects.

While COBRA's 1-step search planning policy is sufficient for our tasks, to solve long-term credit assignment for complicated tasks we would likely need to do multi-step rollouts, such as Model Predictive Coding or Monte-Carlo Tree Search.
One orthogonal but important limitation of stateless rollout-based planning is the inability to solve tasks that require memory, such as a task where a goal location temporarily flashes on the screen.

In summary, natural directions for future work include: more complex environments (e.g. 3 dimensions, physics, and richer visuals); more complex avatars (e.g. embodied, with multiple joints/limbs); learning to set goals; multi-step planning in the exploration phase; and learning a policy at task-time rather than simple search.

\vspace*{0.0in}

\subsubsection*{Acknowledgments}

We would like to thank Matt Botvinick, Tiago Ramalho, Tejas Kulkarni, and Csaba Szepesvari for helpful discussions and insights.

\clearpage

\bibliographystyle{abbrvnat}
\setlength{\bibsep}{5pt} 
\setlength{\bibhang}{10pt}

\bibliography{references}

\newpage

\begin{appendices}\label{S:supp}

\appendix

\renewcommand{\thefigure}{S\arabic{figure}}
\renewcommand{\thetable}{S\arabic{table}}

\section{Agent Videos}\label{S:supp_agent_videos}

Follow this link to see videos of our agent's performance on the training and robustness testing modes of all of our tasks and a README summarizing them:

\url{https://drive.google.com/drive/folders/1JgBPltIB2E8b_RffcvLpEzc50kjj8QBG?usp=sharing}

\section{Model Architectures}\label{S:supp_model}

In all networks we used ReLU activations, weights initialized by a truncated normal \citep{ioffe_2015}, and biases initialized to zero.

\subsection{Vision Module}\label{S:supp_model_vision}

For the vision model we use MONet, with nearly all the same hyperparameters as in \citet{burgess2019}.
The only differences were:
(i) we use a 3-layer Spatial Broadcast Decoder (instead of 4-layer)
(ii) we use $\sigma_{bg} = 0.08$ and $\sigma_{fg} = 0.15$.
These differences seemed to improve disentangling and decomposition slightly for our dataset, though not much --- the default MONet parameters worked well out-of-the-box.

We set the number of entities in the model to 8.
Note that this is more than the maximum number of objects in our exploration phase environment (which is 6), but MONet always uses one slot to encode the background (even though that is black in our case) and an extra slot does not hurt the model.

We preprocess each input image to MONet by rescaling its (R, G, B) color channels the color channels by (1.25, 1.0, 0.75) respectively.
This makes training slightly more stable and efficient by asymmetrizing the color generative factors in the dataset, helping the VAE's color latents emerge more easily.

\subsection{Transition Module}\label{S:supp_model_transition}

Our transition model consists of a slot-wise MLP which outputs both the predicted next scene $\tilde \vz_{t+1}$ and a transition error prediction.
The model itself is a single network $T_{net}$, which is an MLP with 3 hidden layers each of size 512 and an output of size 9.
We treat the first 8 components of this output as the delta of the scene representation and the last component as an error prediction contribution, which we sum over slots to get the error prediction $e_{pred}$.
Specifically, for the scene representation prediction we have $\tilde \vz_{t+1} = \vz_t + T_{net}(\vz_t)[1:8]$ (note that for the sake of clarity we did not mention this in the main text).
We use the error prediction output $e_{pred}$ as a proxy for the transition model's error when training the exploration policy adversarially because it is more efficient and stable to pass gradients through than the full MONet decoder.

The model has 3 loss terms:
\begin{itemize}
    \item Future-prediction loss $L_T = ||V_{dec}(\tilde \vz_{t+1}) - \vx_{t + 1}||^2$.
    \item Error-prediction loss $||e_{pred} - L_T||^2$.
    \item Regularization $|T_{net}(\vz_t)|$ (since $T_{net}$ predicts the delta of the scene representation).
\end{itemize}

The total loss for $T_{net}$ is the sum of these terms (with no reweighting/coefficients).

We believe that predicting the delta of the scene representation is not necessary.
Namely, we expect letting $\tilde \vz_{t+1}$ itself be the output of $T_{net}$ and removing the regularization loss term would work just as well, though have not tried this simplification of the model.

\subsection{Exploration Policy}\label{S:supp_model_exp}

The exploration policy has a single MLP $D_{net}$ with 2 hidden layers each of size 64.
The output size of this MLP is 8, because (while not mentioned in the main text) this network outputs the mean and scale of a 4-dimension Gaussian distribution, from which we sample to get a deformation.
We found that using a distribution like this rather than predicting the deformation deterministically helps training efficiency and stability.

We train this network during the exploration phase.
Given an action $\va_t = \vu + D_{net}(\vz_t, \vu)$ (where $\vu$ is sampled uniformly in $[0, 1]^4$), we use the following 3 loss terms:
\begin{itemize}
    \item Transition model error prediction $e_{pred}(\vz_t, \va_t)$, where the gradients are passed through the transition model without updating the transition model network $T_{net}$ itself.
    \item Deformation regularization $|D_{net}(\vz_t, \vu)|$.
    For this we use a coefficient of $210$.
    \item If the scale of any coordinate of the deformation distribution is less than $0.02$, we add a penalty of $-10^5$ times that coordinate's scale.
    This ensures stability.
    
\end{itemize}

\subsection{Reward Predictor}\label{S:supp_model_reward}

The reward predictor must map a tensor $\vz_t$ of slots representing an environment state to a scalar reward prediction $\tilde r$.
To take advantage of the slot-structured representation of objects in $\vz_t$, we use a Relation Network \citep{santoro2017rn}, a sub-type of graph network \citep{graph_nets}.
Specifically, we use two MLPs $R_{slot\_pair}$ and $R_{global}$ with layer sizes (128, 128) and (128, 1) respectively.
We compute the reward prediction as $\tilde r = R_{global}\left(\sum_{i < j} R_{slot\_pair}\texttt{concat}(\vz_t^i, \vz_t^j)\right)$.

\subsection{Baselines}\label{S:supp_model_baselines}

For all MPO-based models, we used the hyperparameters in Table \ref{table:mpo_hyperparams}.
We used trajectories of length 2 in the replay (longer trajectories yielded lower performance on our tasks, likely because some of the tasks can often be solved in 3 or 4 steps).

For the MPO from slot representations, we used a slot-structured network similar to our agent's reward predictor for both the actor network and the critic network.
Specifically, we applied a per-slot MLP to each slot, then summed across slots, then applied a global MLP.
For the actor network, the per-slot MLP had output sizes (128, 128) and the global MLP had output sizes (128, 128, 8) (the final output size must be 8 since the action space is 4-dimensional).
For the critic network, the per-slot MLP had output sizes (512, 512) and the global MLP had output sizes (256, 1).

For the MPO from image, both the actor and critic had the same architecture: We applied a 2-layer CNN with kernel sizes 3x3, 32 channels per layer, and stride 2 to the input image, then flattened and applied a 3-layer MLP with hidden sizes (256, 256).

These network sizes we chose after numerous hyperparameter sweeps, enough that we can confidently say that neither changing the sizes of any MLP layers by a factor of two nor changing the number of layers in any of these networks improves overall performance.

\begin{table}[h!]
    \begin{center}
    \begin{tabular}{c | c | c | c | c | c}
    $\epsilon$ & $\epsilon_\mu$ & $\epsilon_\Sigma$ & $\gamma$ & target update period & batch size
    \\
    \hline
    0.2 & $10^{-2}$ & $10^{-5}$ & 0.99 & 200 & 512
    \end{tabular}
    \caption{MPO hyperparameters.}
    \label{table:mpo_hyperparams}
    \end{center}
\end{table}

All MPO-based models were trained for $10^6$ gradient steps with batch size 512 using Adam optimizer with learning rate $3\cdot 10^{-4}$.

\section{Training details}\label{S:supp_training}

For our agent, we pre-trained the vision module on a datasets of $2\cdot 10^6$ random static frames from the exploration environment in Spriteworld.
We did not try training it online with the transition model and exploration policy, but because of the modularity of the model believe that would work similarly (as long as the transition model and exploration policy don't get caught in local minima while the vision module is training).

We trained the vision module used the RMSProp optimizer with learning rate $10^{-4}$ as in \citep{burgess2019} and batch size 16 for $7\cdot 10^5$ gradient steps.

We trained the transition model and exploration policy using the Adam optimizer \citep{adam} with learning rate $3\cdot 10^{-4}$ for $5\cdot 10^5$ gradient steps with batch size 16.
The exploration environment was initialized with a random number of objects per episode (between 1 and 7), each with random random shape, color and initial locations.
Exploration environment episodes lasting 10 steps.
For efficiency, we used a distributed actor setup: 1 learner running on GPU trained the transition model and exploration policy networks, while 32 actors runnning on separate CPUs used the exploration policy to collect environment transitions and write them to a replay with capacity $10^5$.
Each actors fetched exploration policy variables from the learner once per 50 environment steps.
Thus our agent was trained off-policy during the exploration phase.

Instead of a uniform replay buffer, we used prioritized replay \citep{schaul2015prioritized}, which sped up training in the unsupervised exploration phase.
For the priority of replay observations we use the error of the transition model.
We set the priority exponent $\alpha=1$ and the importance sampling exponent $\beta=1$.
See \citet{schaul2015prioritized} for details.

The total number of environment steps used by our model's exploration phase is strictly upper-bounded by $10^7$, though due to the distributed nature it is difficult to determine the number precisely.
Note that for COBRA this exploration phase need only be trained once, after which any number of tasks can be learned.
Note also that the MPO handheld baseline also requires this exploration phase to train the vision representations and exploration policy.

For the task-phase, we trained our COBRA agent with branching factor \texttt{B} = 128, training factor \texttt{N} = 10, and \texttt{batch\_size} = 16.
We used an epsilon-greedy training policy with \texttt{epsilon} = 0.2 and trained the agent for 1000 environment episodes.
Depending on the task and how quickly the agent learned, this corresponded to somewhere between 5,000 and 30,000 environment steps.
This was sufficiently long for COBRA to reach what appeared to be asymptotic performance on all the tasks.

All MPO-based models were trained for $10^6$ gradient steps with batch size 512 using the Adam optimizer \citep{adam} with learning rate $3\cdot 10^{-4}$.
This was sufficiently long to reach what appeared to be asymptotic performance on all the tasks.
Like our agent's exploration policy, this used a distributed actor/learner setup with 1 GPU learner and 32 actors.

The distributed-actor nature of both the MPO models and our agent during its exploration phased makes drawing conclusive claims about data efficiency inherently difficult:
While we swept over the number of actors and found a significant degradation of MPO learning efficiency with fewer actors (i.e. when the learner must reuse a greater portion of the replay during training).
However, fully exploring the performance/data efficiency trade-off as the number of actors is varied was beyond our scope.
We do note that the rate of learner gradient steps for all MPO baselines far exceeds the rate of data-collection by the actors (by a factor of ~2 in the vision-based baselines and a factor of ~100 in the state-based baselines), so they were certainly reusing their replay buffer to a large extent.
In fact, on most tasks the replay reuse probability of the state-based MPO models was very similar to the replay reuse probability of our agent at task-time.

Consequently, while the exact data efficiency numbers we report should be taken with a grain of salt, we're confident that the data efficiency difference between our model and the baselines is very significant (given that it is several orders of magnitude).

\section{Model Variations Considered}\label{S:supp_things_tried}

Here we recount some of the model variations we explored before arriving at the model described in the main text.
We focus on the transition model and exploration policy, which are the primary algorithmic contributions of this work.
We hope this may be useful for researchers exploring closely related models.

\subsection*{Transition Model}

Before converging on the pixel-loss-via-decoder method described in Section \ref{S:model}, we tried a variety of methods to train the transition model with a loss in latent space.
The aim here was to pressure $T(\vz_t, \va_t)$ to be similar to $\vz_{t+1}$.

However, as mentioned in the main text, the slot ordering of $\vz_{t+1}$ may differ from that of $\vz_t$, because the ordering that MONet encodes the slots is a complicated implicit function that is sensitive to small changes in the visual input.
Thus to compute a loss in latent space we need to solve a matching problem.

There are many ways to find such a matching just by looking at the slot representations and without considering all possible slot orderings.
One that worked well for us was to take each slot of $\vz_{t+1}$ and match it with the slot of $\vz_t$ with which it had lowest mean squared error (allowing for double-assignment).
Using KL divergence instead of mean squared error also worked.

However, we were bothered by a subtle drawback of these matching methods:
They are based purely on static scenes across time without considering how objects move.
For example, if two similar-looking objects cross paths and seem to ``switch places'' in consecutive timesteps, they could be incorrectly matched, which would throw off the transition model's training.

Thus we find such matching approaches unnatural.
We prefer using pixel loss through the MONet decoder, as it is a very general principle and uses simplicity of dynamic prediction to determine the slot-ordering of future timesteps.

\subsection*{Exploration Policy}

As mentioned in Section \ref{S:discussion}, our exploration policy is a form of distributional RL.

The first approach we tried for the exploration policy was to parameterize the action distribution density function by a neural network and sample actions via rejection sampling.
This is effectively the method used in \citep{haber2018learning} and works when the curiosity signals are dense.
However, we found the sampling to be computationally expensive in our setting of sparse curiosity signals, and were intrigued by the prospect of learning a way to sample direction without rejection, hence quantile regression.

As mentioned in Section \ref{S:discussion}, we initially tried using a multi-dimensional version of the Huber quantile function loss typically used in distributional RL, but this was very slow to train.
Controlled experiments with artificial curiosity functions showed it to scale poorly both with sparsity of the curiosity function and with dimensionality of the action space.

We then arrived at our deformation method, which trained quickly and scaled surprisingly well:
In experiments with artificial curiosity signals we found it to work well even in 50-dimensional action spaces.
In addition to working well in our context, we find the prospect of learning attractor dynamics in an action space quite appealing, and imagine similar approaches may be more generally useful in continuous control.

\subsection*{Curiosity Signal}

As mentioned in Section \ref{S:discussion}, our use of the transition model error as a curiosity signal is rudimentary, though could easily be extended to a more general method as in \citep{pathak2017curiosity}.

However, we originally hoped to use an entirely different method altogether: We hoped to use the intrinsic uncertainty of the transition model as a metric of curiosity.
Namely, we aimed to build a stochastic transition model and use its variance as curiosity.
We came very close to getting this to work, but were ultimately foiled by some subtleties about the MONet's representation.

Specifically, making the transition model stochastic was easy --- that can be done by letting the transition model parameterize the mean and variance of the slots in $\tilde \vz_{t+1}$. It can be trained via a KL penalty with the distribution of $\vz_{t+1}$ if using a slot-matching loss, or if using the decoder-pass-through pixel loss can be trained by sampling from $\tilde \vz_{t+1}$ before decoding.

The trouble comes when trying to use the variance of the transition model as uncertainty.
As shown in Figure \ref{fig:scene_action_transition}, in order to handle a variable number of objects MONet lets some of its slots encode no object (e.g. a blank image).
This implies that there is a region in latent space that is decoded to a blank image.
Consequently, when MONet infers a blank slot it can afford to use high variance in the latent representation.
In practice, the variance for blank slot encodings in MONet is much higher than that for non-blank slots, and this is amplified with the number of blank slots (which, given our exploration environment had 1-6 objects and 8 slots, was more than 50\%).
Hence the high variance for blank slots in the stochastic transition model drowns out the signal from uncertainty about objects moving, so is a poor curiosity signal.

There are certainly some tricks one could use to circumvent this problem (e.g. explicitly excluding blank slots from the uncertainty calculation, or dividing by inferred variance).
However, we found these unsatisfactory and did not see an elegant solution so ended up using the pixel loss, which is very simple but works.

\section{Environment details}\label{S:supp_environment}

Our Spriteworld environment was a square, 2-dimensional world with geometric objects varying in position, shape, and color.
We rendered this environment into images of shape (64, 64, 3) with an anti-aliasing factor of 5.
While the open-sourced version of Spriteworld (\url{https://github.com/deepmind/spriteworld/}) uses a PIL-based renderer, for the earlier version of Spriteworld used in this paper we instead used a PyGame-based renderer (though this choice of renderer has no effect on the results).
Objects could occlude in a consistent manner with each object living on a different z-layer but could not collide/interact in any other way.
Object size did not vary in the environment.
Instead, each shape's size was fixed so that its area was $0.017$ in units of squared frame-width.

As mentioned in the main text, the action space for Spriteworld is the continuous hypercube $[0, 1]^4$.
We view the first two components of an action as a "position click" $(x, y)$ and the second two as a "motion click" $(\delta x, \delta y)$.
If the position click, when viewed as point in the rendered environment, lands within the boundary of an object, that object is moved in the direction of the motion click (centered so that a motion click of $(0.5, 0.5)$ corresponds to zero motion).
The magnitude of the motion is scaled by a factor of 0.25 and we add scale-0.05 Gaussian noise to motions, so $(x_{t+1}, y_{t+1}) = (x_t, y_t) + 0.25 \cdot (\delta x, \delta y) + \mathcal{N}(0, 0.0025)$.
Note that because the motion click is centered, $\delta x$ and $\delta y$ are both in $[-0.5, 0.5]$, hence the furthest an object can move in each direction is 0.125 (and the furthest overall is $0.125 \cdot \sqrt{2} = 0.177$), so the agent can move an object from any point in the environment to any other point within 8 steps.
Note also that including the Gaussian noise added to the motion did help the transition model/exploration policy training loop during our agent's exploration phase, as it introduced noise to object motions (hence unavoidable prediction error when objects moved).
However, even without this noise the exploration policy did learn to click on objects though showed some biases to click near object borders.

We ensure that objects never exit the frame.
To do this, after an object moves we clip each coordinate of it's position to $[0, 1]$.

For each task, the Spriteworld environment procedurally generated each episode from an initialization distribution.
Table \ref{table:task_specs} summarizes some of the properties of the tasks.
The details not easily tabulated are detailed as follows:

\begin{itemize}
    \item \textbf{Shape Robustness}.
    During training the object's shape is a square.
    During robustness testing it is in {circle, triangle}.
    We used a shaped reward which is linearly inversely proportional to the distance from the object to the goal location, which is always the center of the frame.
    If the object is brought within a distance 0.075 (in units of the frame width) of the center of the frame, the episode is considered a success and resets (assuming this happens before the maximum episode length timeout).
    
    \item \textbf{Position Robustness}
    One object with hue in $[0.0, 0.4]$ is the target.
    One object with hue in $[0.5, 0.9]$ is the task-irrelevant distractor.
    During training, the target's position is initialized randomly uniformly in the frame except the lower-right quadrant.
    For robustness testing, the target's position is initialized randomly uniformly in the lower-right quadrant.
    The distractor's position is unconstrained in $([0, 1], [0, 1])$ for both modes.
    The reward and termination criterion is the same as in the shape robustness task as a function of the target only.
    
    \item \textbf{\# Targets Robustness}
    The target(s) has/have shape square.
    The distractors have shape in {circle, triangle}.
    During training there is 1 target and 2 distractors.
    During robustness testing there are 2 targets and 2 distractors.
    Again, the reward is linearly inversely proportional to the distance from the target to the center of the frame.
    When there are two targets, the reward is the sum of the reward-per-target of each.
    When there are two targets, the episode terminates only when both meet the termination criterion, i.e. both are within a distance 0.075 of the frame center.
    
    \item \textbf{\# Distractors Robustness}
    The targets have shape square.
    The distractor(s) has/have shape in {circle, triangle}.
    During training there are 2 targets and 1 distractors.
    During robustness testing there are 2 targets and 2 distractors.
    Again, the reward is the sum over targets of their goal-finding rewards.
    
    \item \textbf{Sorting}
    There are 5 narrow hue distributions:
    red [0.9, 1], blue [0.55, 0.65], green [0.27, 0.37], purple [0.73, 0.83], and yellow [0.12, 0.22].
    Each color has it's own goal location, which are [0.75, 0.75], [0.75, 0.25], [0.25, 0.75], [0.25, 0.25], and [0.5, 0.5] respectively.
    During training a random pair of colors is sampled, except not the pair (red, blue).
    Objects of these colors are generated, and the agent must bring each to their respective goal location.
    For robustness testing, the held-out (red, blue) color pair.
    This tests robustness to an unseen combination of objects and the agent's ability to correctly factor and recompose the objects' respective goals.
    The reward here is the sum of the per-object goal-finding rewards of the two objects in each episode, except here the goal locations are not the center of the frame for all colors except yellow.
    Agents are also given a bonus reward if it successfully completes the task.
    this did not affect performance of our agent at all, but did help the MPO baselines.
    
    \item \textbf{Clustering}
    Given the color hue distributions in the sorting task, the environment samples a pair of colors for each episode and generates 2 objects of each color.
    The agent is rewarded for clustering the objects by color, namely bringing each pair of similarly-colored objects together while creating a sufficient inter-pair distance.
    During training, the color pair is always (blue, green).
    For robustness testing, the color pair is (purple, yellow).
    To compute the reward we used the inverse of the Davies-Bouldin clustering metric  \citep{davies_bouldin} and terminated the episode when this inverse clustering metric is higher than 2.5, in which case the agent received a bonus reward (which, as in the sorting task, helped only the MPO baselines).
    
\end{itemize}

\begin{table*}[t]
    \centering
    \begin{tabular}{c | c | c | c | c | c |}
    Task &
    \makecell{maximum \\ episode \\ length} &
    \makecell{color \\ (H, S, V)} &
    \makecell{initial position \\ (x, y)} &
    shape &
    \makecell{number of \\ sprites \\ (training)}
    \\
    \hline
    exploration
    & 2 & $([0, 1], [0.3, 1], 1)$ & ([0, 1], [0, 1]) & [square, circle, triangle] & [1, 6]
    
    \\
    \hline
    \makecell{goal\_finding \\ shape robustness}
    & 20 & $([0, 0.4], [0.3, 1], 1)$ & ([0, 1], [0, 1]) & see text & 1
    
    \\
    \hline
    \makecell{goal\_finding \\ init position robustness}
    & 20 & see text & see text & [square, circle, triangle] & 2
    
    \\
    \hline
    \makecell{goal\_finding \\ \# targets robustness}
    & 20 & ([0, 0.5], [0.3, 1], 1) & ([0, 1], [0, 1]) & see text & 3
    
    \\
    \hline
    \makecell{goal\_finding \\ \# distractors robustness}
    & 20 & see text & ([0, 1], [0, 1]) & see text & 3
    
    \\
    \hline
    sorting
    & 50 & ([0, 0.5], [0.3, 1], 1) & ([0, 1], [0, 1]) & [square, circle, triangle] & 5
    
    \\
    \hline
    clustering
    & 50 & see text & ([0, 1], [0, 1]) & [square, circle, triangle] & 4

    \end{tabular}
    \caption{Specifications for each task. These are the easily-summarized components of the task specifications. See text for the boxes that are more complicated.}
    \label{table:task_specs}
\end{table*}

\section{Sparse rewards}\label{S:supp_sparse_rewards}

\begin{figure}[t!]
  \centering
  \includegraphics[width=0.6\linewidth]{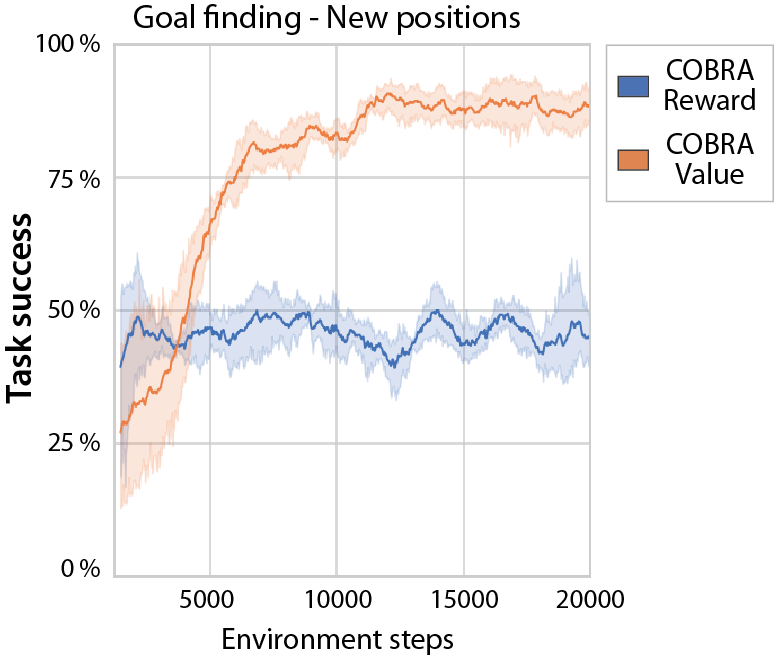}
  \caption{
  \textbf{Training curve in Sparse reward situation}\\
  Comparison between a COBRA agent using a Reward predictor or a Value predictor, when solving Goal Finding tasks with sparse terminal rewards only.
  Moving average over 50 episodes, shaded color indicates one standard deviation around the median, over 5 replica.
  }\label{fig:supp_sparse_reward}
\end{figure}

In the tasks we presented so far, we used dense, shaped rewards.
This meant that a single step search algorithm and a reward predictor are sufficient to perform optimally.
However, we could relax this requirement, and use sparse rewards instead.
This requires deeper search rollouts, or the use of a value function.

We implemented a Value function version of our search agent, where the Reward predictor is replaced by a Value predictor.
This can be trained in a similar fashion, using TD-learning:
\begin{align}
    \text{Given} &\text{ replay transitions } (\vx_t, \va_t, \vx_{t+1}, r_t, \gamma_t) \nonumber\\
    \vz_t &= V_{enc}(\vx_t), \vz_{t+1} = V_{enc}(\vx_{t+1}) \nonumber\\
    \tilde{\vz}_{t+1} &= T(\vz_t, \va_t) \nonumber\\
    \text{Loss}_{\text{TD}} &= || r_t + \gamma_t V(\vz_{t+1}) - V(\vz_t) ||^2\\
    \text{Loss}_{\text{TD predicted}} &= || r_t + \gamma_t V(\tilde{\vz}_{t+1}) - V(\vz_t) ||^2\\
    \text{Loss}_{\text{consistency}} &= || V(\vz_{t+1}) - V(\tilde{\vz}_{t+1}) ||^2
\end{align}
We found that using a sum of these three losses offered the fastest and most stable estimation of the Value function in our setting, however in some situations we also had good success with using only $\text{Loss}_{TD}$.
Finally, we act so that actions which maximize $V(\tilde{\vz}_t+1)$ are selected with our 1-step search.

We demonstrate this on a modified version of the Goal Finding - Shape robustness task, where we only provide a reward of 1 when the target reached the goal, and 0 everywhere else.
As can be expected, this increases the difficulty of the tasks dramatically, see training curves in Figure~\ref{fig:supp_sparse_reward}.
The Reward predictor version of COBRA only reaches 50\% success, as it can bring sprites appearing close to the target to the goal.
This is because some actions will lead to predicted scenes where the target is close to the center, and due to the smoothness of our reward predictor, there is a non-zero reward signal to follow.
Using a value predictor instead successfully solves all situations, without this requirement for the sprite to being close to the goal.

\section{Vision Module Disentangling}\label{S:supp_disentangling}

\begin{figure}[t!]
  \centering
  \includegraphics[width=0.7\linewidth]{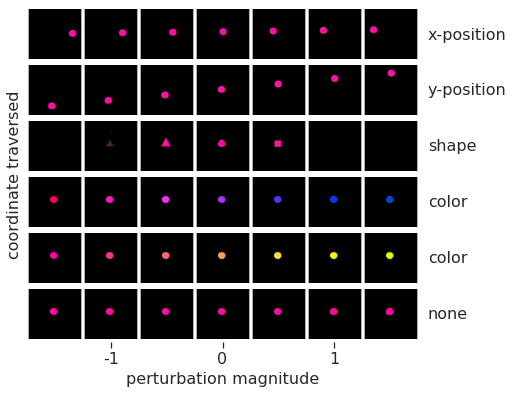}
  \caption{\textbf{Vision module disentangling.}
  Each row show the effect on reconstructions of sweeping one latent component from -1.5 to 1.5 (keeping all other latent components fixed).
  Only the 6 most significant latents are shown --- the model learns two more non-coding "none" latents that revert to its Gaussian prior.
  The latent space is disentangled, as indicated by the labels (assigned post-hoc by eye) on the right.
  }\label{fig:supp_disentangling}
\end{figure}

As shown in \citet{burgess2019}, the MONet model not only learns a scene decomposition but can also learn a disentangled latent space for object representations.
We found that indeed this was the case for our dataset, as can be seen from the latent-space traversals from the model shown in Figure \ref{fig:supp_disentangling}.
We believe that this disentangling may have helped our agent be robust to some perturbations (e.g. perturbing irrelevant features), though we did not explicitly incorporate any attention mechanisms in our reward predictor or encourage it in any way to be invariant to any latent components.

\section{Ablations and Additional Baselines}\label{S:supp_ablation_baselines}

\begin{figure*}[t!]
  \centering
  \includegraphics[width=0.48\linewidth]{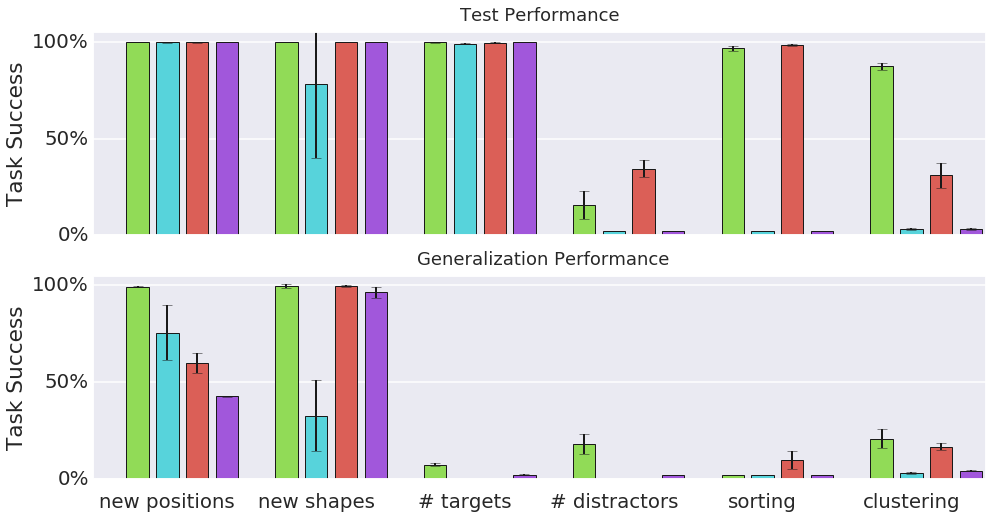}
  \includegraphics[width=0.48\linewidth]{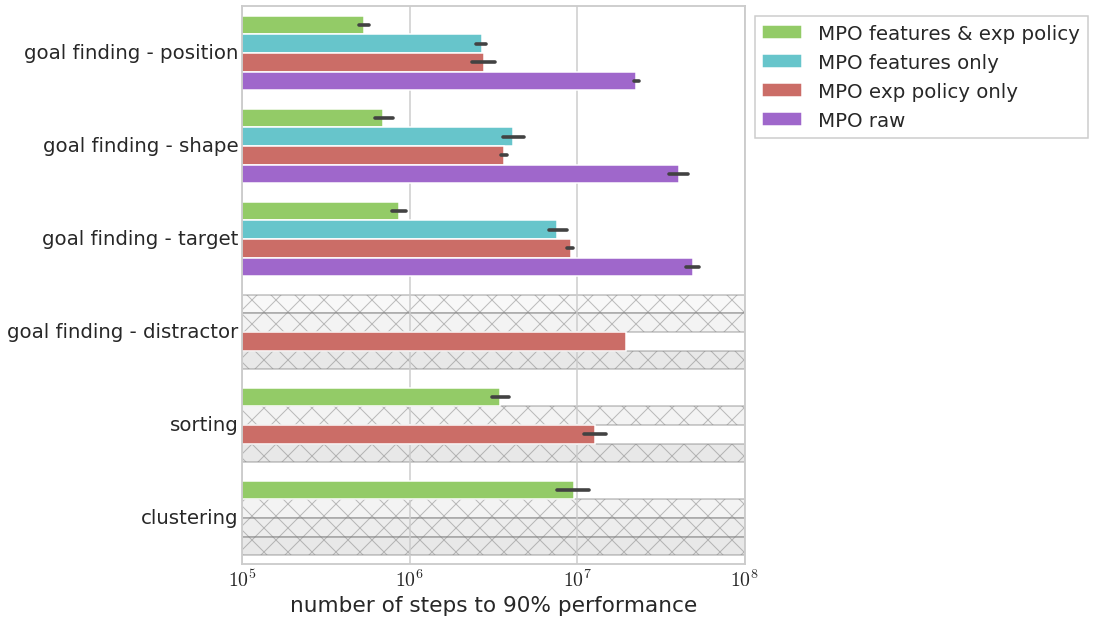}
  \caption{
  \textbf{Ablation study for MPO baselines}\\
  \textit{(left)} \textbf{Performance and Robustness.}
  Test-time agent performance on environments sampled from the training distribution (top) and from the robustness generalization tasks (bottom), analogous to main text Figure \ref{fig:agent_results}-A.
  \textit{(right)} \textbf{Data efficiency.}
  Number of environment steps to reach 90\% test performance on the training task distributions.
  This shows MPO agents with ech combinations of visual slot-structured features (from our model) and exploration policy (from our model).
  The one with both features and exploration policy is what we refer to in the main text as "MPO handheld".
  Note the log scale and axis range of the data efficiency plot.
  }\label{fig:supp_mpo}
\end{figure*}

In the main text we compare our COBRA agent to two baselines: (i) Raw MPO from pixels and (ii) MPO with slot/object-structured MONet features and our agent's learned exploration policy.
While the combination of MONet features and exploration policy improved MPO's performance, it is unclear which factor was primarily responsible for this improvement.
Consequently, here we analyze them independently, running both an MPO with only MONet features and an MPO with only our agent's exploration policy.
These results are shown in Figure \ref{fig:supp_mpo}.
In terms of data efficiency, both of these perform in between the raw MPO and the MPO with both MONet features and exploration policy.
In terms of asymptotic performance and robustness/generalization, it seems that the exploration policy provides a bigger boost than MONet features on our task suite.

\begin{figure*}[t!]
  \centering
  \includegraphics[width=0.48\linewidth]{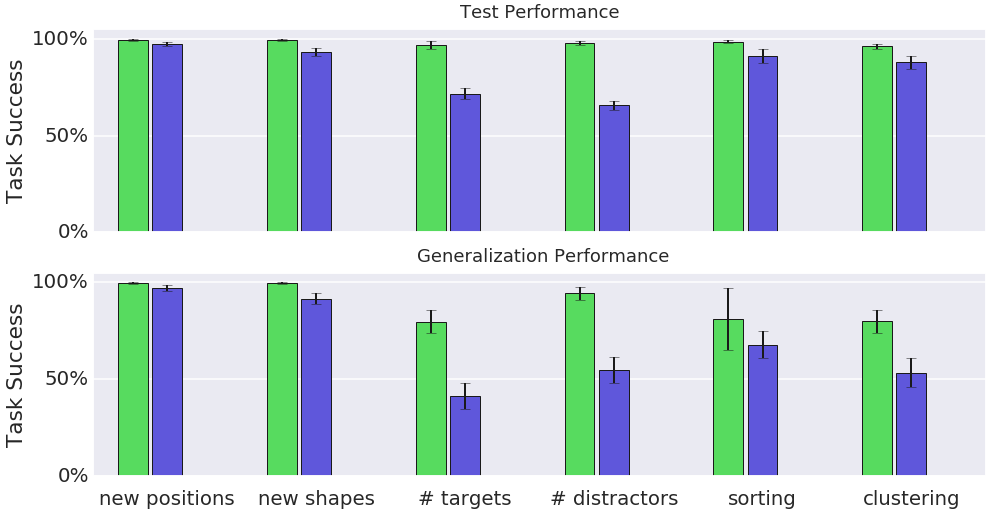}
  \includegraphics[width=0.48\linewidth]{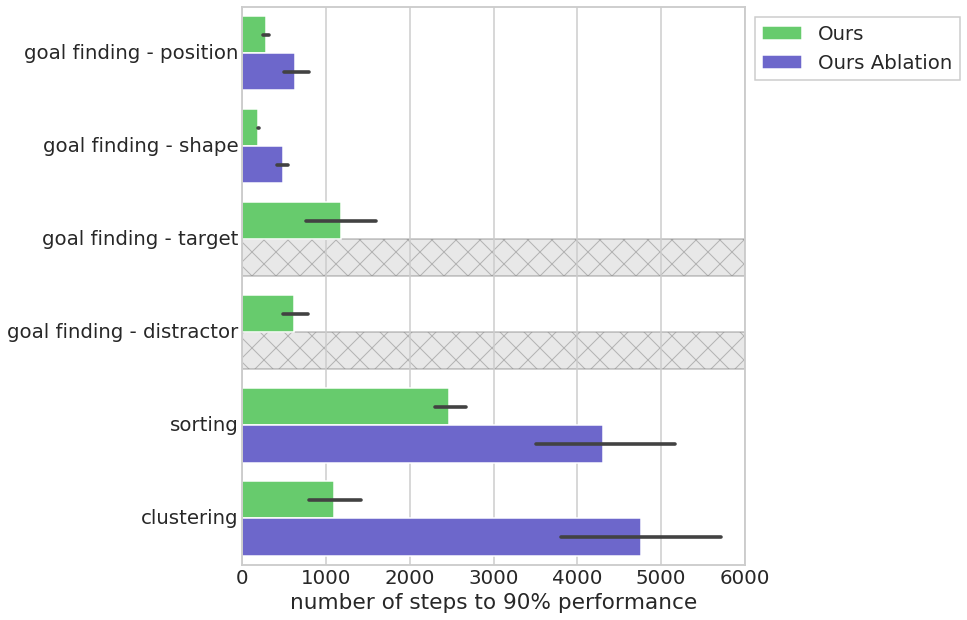}
  \caption{
  \textbf{Ablation of Exploration policy for our agent}\\
  Our COBRA agent is shown in green, and the same agent without using the exploration policy at task time is shown in blue.
  Without the exploration policy, this ablated agent samples actions randomly uniformly in $[0, 1]^4$ when doing model-based search.
  It does, however, use the exploration during the exploration phase (this is always necessary to learn a good transition model).
  Note that unlike Figures \ref{fig:agent_results} and \ref{fig:supp_mpo}, the data efficiency plot here does not use a log scale.
  \textit{(left)} \textbf{Robustness.}
  Test-time agent performance on environments sampled from the training distribution (top) and from the robustness generalization tasks (bottom), analogous to main text Figure \ref{fig:agent_results}-A.
  \textit{(right)} \textbf{Data efficiency.}
  Number of environment steps to reach 90\% test performance on the training task distributions.
  }\label{fig:supp_ablation}
\end{figure*}

\section{Additional Transition Model Rollouts}\label{S:supp_transition_model}

\begin{figure}[t!]
  \centering
  \includegraphics[width=0.35\linewidth]{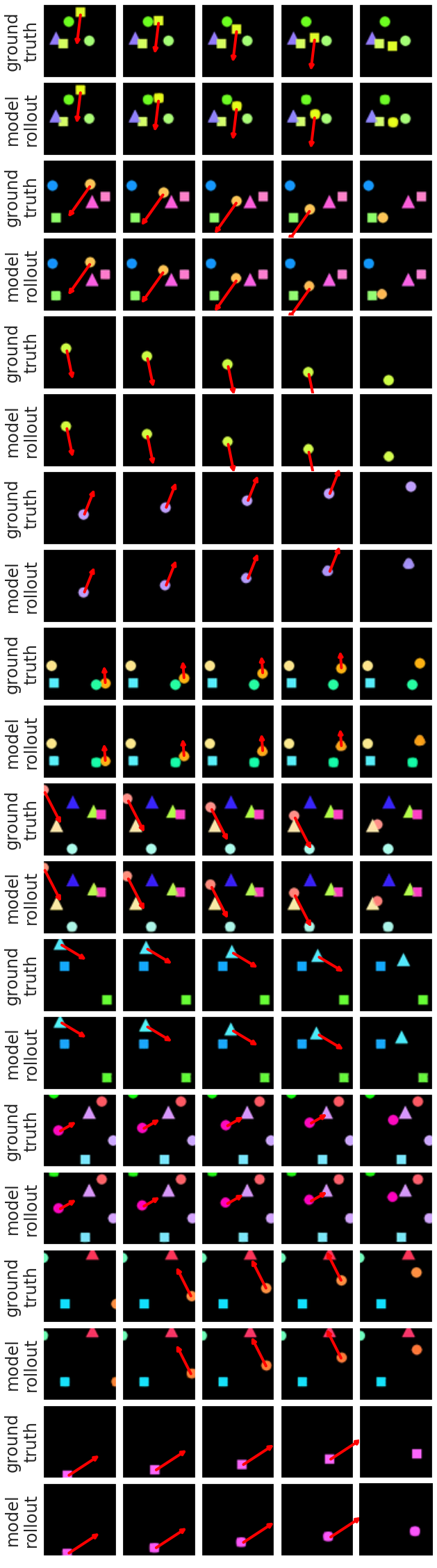}
  \caption{\textbf{Transition model rollouts.}
  As in Figure \ref{fig:scene_action_transition}, each pair of rows compares the ground truth environment observations from a sequence of actions to rollouts from the transition model on the same sequence of actions (using the vision module decoder as a renderer).
  The actions are indicated by the red arrows (which display the motion click vector centered at the position click location).
  These action sequences were generated by using the same motion click repeatedly while moving the position click to the ground-truth object location at each step.
  }\label{fig:supp_transition_model_rollouts}
\end{figure}

In Figure \ref{fig:scene_action_transition} in the main text we show 2 examples of transition model rollouts.
Here in Figure \ref{fig:supp_transition_model_rollouts} we show many more.

\end{appendices}

\end{document}